# A critical analysis of metrics used for measuring progress in artificial intelligence

Kathrin Blagec [1], Georg Dorffner [1], Milad Moradi [1], Matthias Samwald [1]

[1] Section for Artificial Intelligence and Decision Support; Center for Medical Statistics, Informatics, and Intelligent Systems; Medical University of Vienna, Vienna, Austria.

Corresponding author: Matthias Samwald (matthias.samwald [at] meduniwien.ac.at)

---

## Abstract

Comparing model performances on benchmark datasets is an integral part of measuring and driving progress in artificial intelligence. A model's performance on a benchmark dataset is commonly assessed based on a single or a small set of performance metrics. While this enables quick comparisons, it may entail the risk of inadequately reflecting model performance if the metric does not sufficiently cover all performance characteristics. It is unknown to what extent this might impact benchmarking efforts.

To address this question, we analysed the current landscape of performance metrics based on data covering 3867 machine learning model performance results from the open repository 'Papers with Code'. Our results suggest that the large majority of metrics currently used have properties that may result in an inadequate reflection of a models' performance. While alternative metrics that address problematic properties have been proposed, they are currently rarely used.

Furthermore, we describe ambiguities in reported metrics, which may lead to difficulties in interpreting and comparing model performances.

**Keywords:** Artificial intelligence, Machine learning, Performance metrics, Benchmarking

# Introduction

Benchmarking, i.e. the process of measuring and comparing model performance on a specific task or set of tasks, is an important driver of progress in artificial intelligence research. The specific benchmark task and performance metrics associated with it can be seen as an operationalisation of a more abstract, general problem which the research community aims to solve.

Benchmark datasets are conceptualized as fixed sets of data that are manually, semi-automatically or automatically generated to form a representative sample for these specific tasks to be solved by a model.

In the last few years, many benchmark datasets covering a vast spectrum of artificial intelligence (AI) tasks – such as machine translation, object detection or question-answering – across many application domains have been established. (Marcus 1993; Phillips et al 1998; Deng et al 2009; Lin et al 2014; Hermann et al 2015; Rajpurkar et al 2016; Cordts et al 2016) In parallel, spurred by advancements in computational capacities, there has been an increase in the development and publication of models that continuously improve state-of-the-art results on these benchmarks. Projects such as Paper With Code[1] compile overview statistics on progress on benchmarks.

A model's performance on a benchmark dataset is most commonly measured by one single, or more rarely, by a small set of performance metrics. Describing the capabilities of a model through a single metric enables quick and simple comparison of different models.

However, condensing the performance characteristics of a model into a single metric also entails the risk of providing only one projection of model performance and errors, thereby emphasizing certain aspects of these characteristics over others. (Ferri et al 2009; Monteiro et al 2019)

Discussions about the suitability, informative value and weaknesses of prevailing performance measures have a long history. These discussions are driven by the desire to find measures that perfectly describe model capabilities in different domains. New measures are frequently proposed to overcome shortcomings of existing measures, commonly with regard to special use cases.

A well-known example for a problematic performance metric used historically for classification tasks in machine learning is accuracy, i.e. the ratio of correctly predicted samples to the total number of samples. While these shortcomings are now widely recognized—e.g. its inadequate reflection of a classifiers' performance when used on unbalanced datasets—it still continues to be used as a single metric to report model performance.

---

[1] https://www.paperswithcode.com/

Due to such justified criticism, the research community has called for replacing or extending the reporting of accuracy with more informative classification performance measures, such as precision, recall or the F1 score, which combines precision and recall into one single score. For these metrics, in turn, several scenarios have been identified in which they may provide misleading assessments of a classifier's performance. (Chicco and Jurman 2020)

Furthermore, in an experimental analysis based on 30 benchmark datasets Ferri et al. showed that correlations between performance metrics for classification tasks tend to be very low when dealing with imbalanced datasets, which is a frequent situation. (Ferri et al 2009)

Moreover, there exists extensive criticism on the use of accuracy and related discontinuous metrics in general, even with optimally balanced datasets, as they do not fulfill the criteria of proper scoring rules. Proper scoring rules, such as the Brier score or the logarithmic scoring rule, are considered to incentivise transparent prediction and maximization of the expected reward. Proper scoring rules further encourage a clear separation between the statistical component of a prediction problem (i.e., determining the predicted probabilities for each class and for each sample) and the decision component (i.e., defining the threshold for assigning classes based on predicted probabilities). The Brier score can further be decomposed into two terms representing discrimination and calibration (i.e. the assessment of absolute prediction accuracy) components, which provides additional information on the classifiers' performance. (Harrell 2001; Gneiting and Raftery 2007)

Despite these vigorous and multifaceted discussions, it is currently still unclear to what extent efforts to promote metrics adequate for benchmark tasks have made an impact on global AI benchmarking efforts, and where current shortcomings can be identified.

In this analysis we aim to address this question by providing a comprehensive overview of the current landscape of performance measures used by benchmark datasets to track progress in artificial intelligence. To do so, we draw data on 3867 machine learning model performance results reported in arXiv submissions and peer-reviewed journals from the database 'Papers with Code' (PWC) [2], manually curate utilized performance measures, and provide a global view on performance measures used to quantify progress in artificial intelligence.

---

[2] https://paperswithcode.com/

# Results

## Descriptive statistics of the dataset

32209 benchmark results across 2298 distinct benchmark datasets reported in a total number of 3867 papers were included in this analysis. Included papers consist of papers in the PWC database that were annotated with at least one performance metric as of June 2020. (Table 1). A single paper can thus contribute results to more than one benchmark and to one or more performance metrics.

*Table 1: General descriptives of the analyzed dataset (as of 13.07.2020).*

| Number of papers | 3883 |
| --- | --- |
| Number of benchmark results | 32209 |
| Number of benchmark datasets | 2298 |
| Time span of manuscripts covered by the dataset | 2000-2020 |

The publication period of the analyzed papers covers twenty years, from 2000 until 2020, with the majority having been published in the past ten years (Figure 1).

Figure 2 shows the number of benchmark datasets per high-level process. 'Vision process', 'Natural language processing' and 'Fundamental AI process', which includes general methods such as transfer and meta learning, were the three processes with the highest number of associated benchmark datasets.

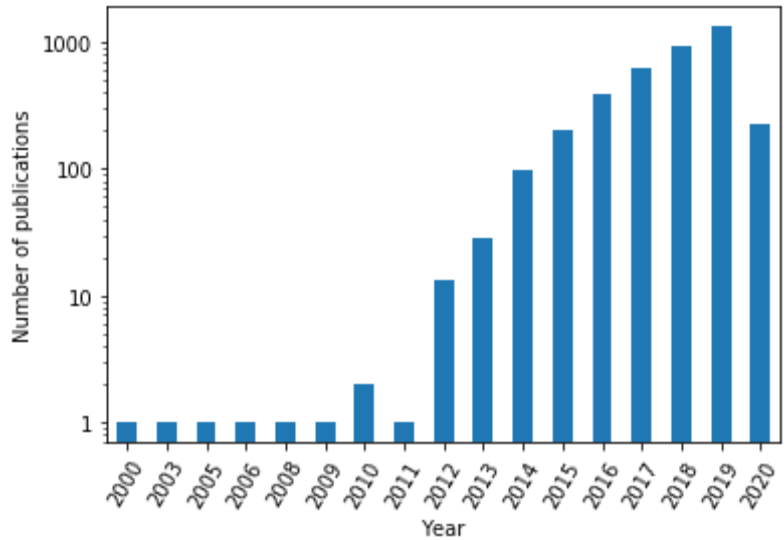

*Figure 1: Number of publications covered by the dataset per year. The y-axis is scaled logarithmically.*

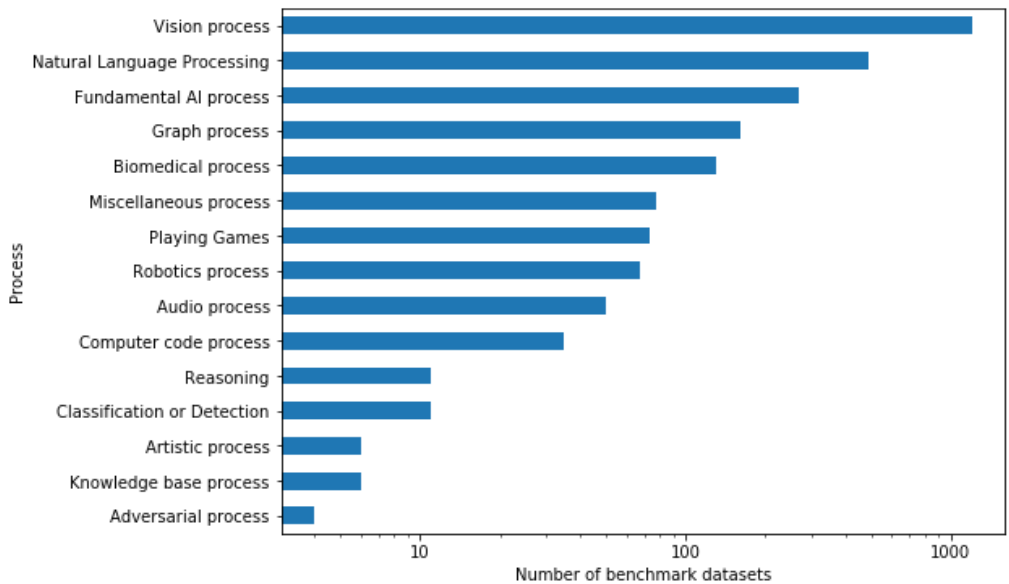

*Figure 2: Number of benchmark datasets per higher level process. A benchmark dataset can be mapped to more than one process category. The x-axis is scaled logarithmically.*

## Most frequently reported performance metrics

The raw dataset exported from PWC contained a total number of 812 different metric names that were used by human annotators to add results for a given model to the evaluation table of the relevant benchmark dataset's leaderboard.

We conducted extensive manual curation of this raw list of metrics to map performance metrics into a canonical hierarchy. Thereby we dealt with several complexities in which performance metrics are reported, such as: (1) The same metric may be reported under many different synonyms and abbreviations in different research papers. For example, 63 naming variations of the F1 score appeared throughout the dataset, such as 'F-Measure', 'H-Mean' and 'F1'. These were all mapped to a single, canonical property. (2) Some performance metrics are reported in different sub-variants. For example, the performance metrics 'hits@1', 'hits@3' and 'hits@10' were considered as sub-metrics of the top-level metric hits@k. Throughout the paper, we will refer to canonical properties and mapped metrics as 'top-level metrics' and 'sub-metrics', respectively. Further examples and a description of the rationale behind the mappings are provided in the 'Methods' section.

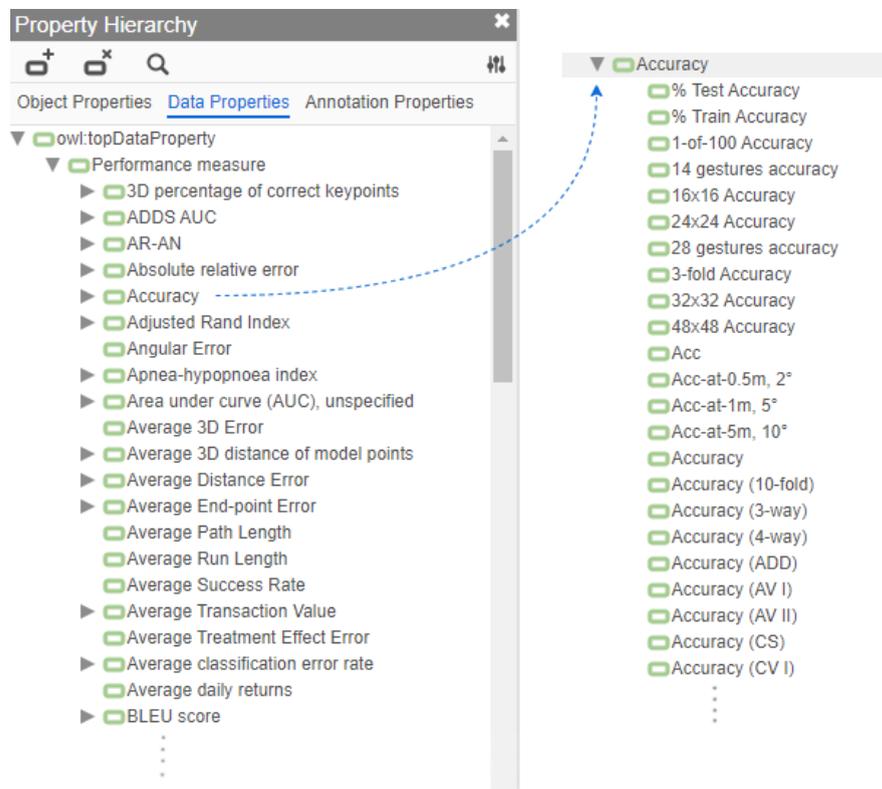

*Figure 3: Property hierarchy after manual curation of the raw list of metrics. The left side of the image shows an excerpt of the list of top-level performance metrics; the right side shows an excerpt of the list of submetrics for the top-level metric 'Accuracy'.*

After manual curation, the resulting list covered by our dataset could be reduced from 812 to 187 distinct top-level performance metrics. 271 entries from the original list, such as 'All' or 'Test' could not be assigned to a metric and were subsumed under a separate category 'Undetermined'. Figure 3 shows an excerpt of the curated property hierarchy.

Top-level metrics were further categorized based on the task types they are usually applied to, e.g. 'accuracy' was mapped to 'classification', 'mean squared error' was mapped to 'regression' and 'BLEU' was mapped to 'natural language processing'. The distribution of metric types and the number of benchmark datasets associated with these metric types is shown in Figure 4. Classification metrics were, by far, the metric type with the highest number of associated benchmark datasets.

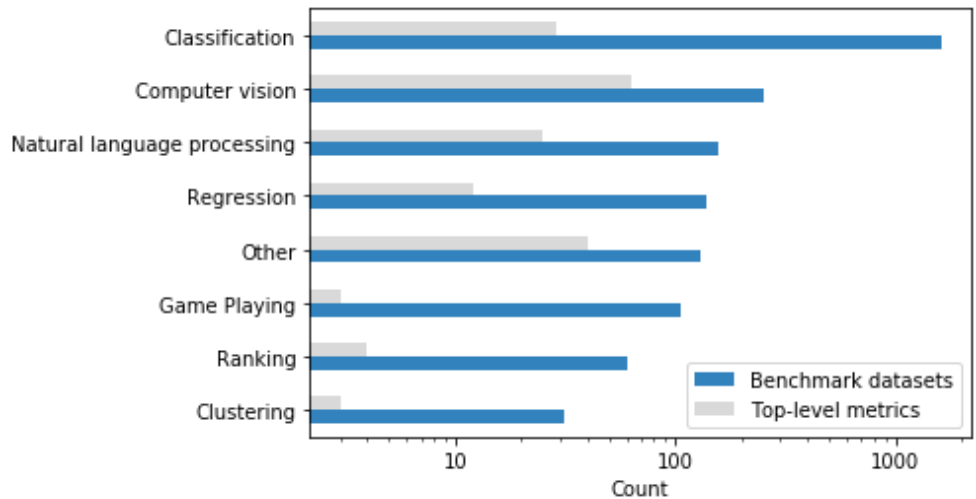

*Figure 4: Number of top-level metrics per metric type (blue bars) and number of distinct benchmark datasets that use at least one top-level metric of the respective metric type (grey bars). 'Other' includes for example the sub-metric 'Resource requirements' which is used to compare meta-level model information, such as the number of model parameters. A benchmark dataset can be mapped to more than one metric type. The x-axis is scaled logarithmically.*

Figure 5 shows the ten most frequently reported performance metrics in terms of the number of benchmark datasets that use this metric to evaluate model performance overall and per task type. Complete statistics for all 189 metrics covered by the dataset are made available online (see Section 'Data and code availability').

*Accuracy* was by far the most frequently used performance metric, being used by 38% of all benchmark datasets covered in this analysis. The second and third most commonly reported metrics were precision and the F-measure with 16% and 13% of all benchmark datasets using them to evaluate model results. Considering submetrics, F1 score was the most frequently used F-measure.

Examples for well-established but infrequently used metrics throughout the analyzed dataset include specificity, with only one occurrence, and metrics for regression tasks, such as mean squared and absolute error, root mean square deviation and $R^2$, which appeared rarely in the analyzed dataset, being used, taken together, by only 5% of benchmark datasets.

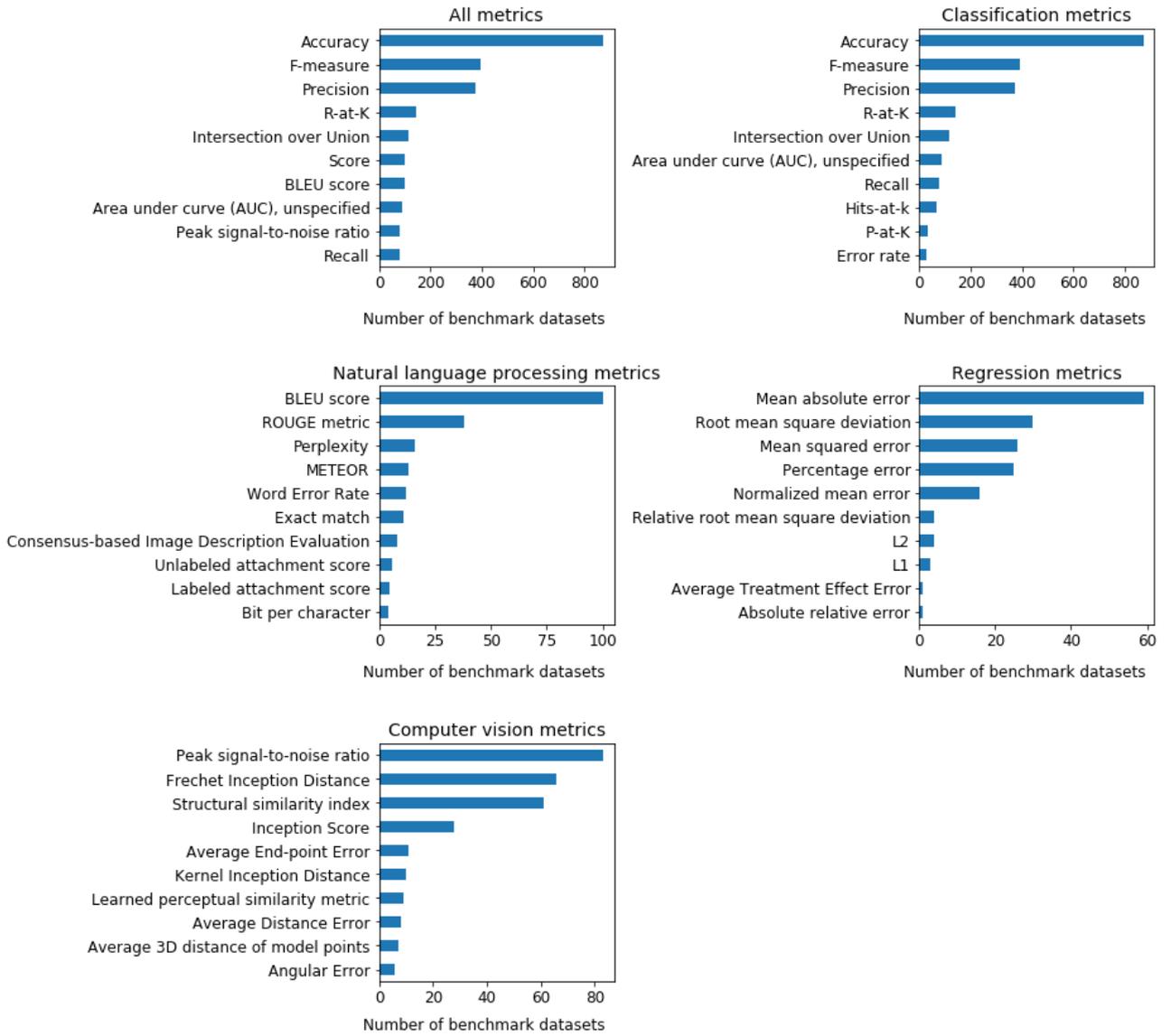

**Figure 5:** Top 10 most frequently reported performance metrics, overall and per metric type. The x-axis shows the number of distinct benchmark datasets that are evaluated using the respective performance metric. A single benchmark dataset can contribute to multiple metrics counts. 'Area under curve (AUC), unspecified': AUC metrics that were not further specified by the annotators.

## Performance metrics per benchmark datasets and co-occurrence of performance metrics

For more than two thirds (77.2%) of the analyzed benchmark datasets, only a single performance metric was reported (see Figure 4), when considering only top-level metrics. 14.4% of the benchmark datasets had two distinct annotated top-level metrics and 6% had three distinct annotated top-level metrics.

The minimum and maximum count of distinct top-level metrics per benchmark dataset were 1 and 6, respectively, and the median number of distinct top-level metrics was 1.

In 83.1% of the benchmark datasets where the top-level metric *accuracy* was reported, no other top-level metrics were reported. In 60.9% of the benchmark datasets where the *F-Measure* was reported, no other top-level metrics were reported.

When considering sub-metrics, the statistics slightly change. The proportion of benchmark datasets that report only a single performance metric decreases to 70.4%, while the fraction of benchmark datasets that use two or more metrics, slightly increases (see Figure 6). This increase when considering sub-metrics can be explained by the reporting of different sub-variants of a given metric. For example, the reporting of 'hits@1', 'hits@3' and 'hits@10', which are all sub-variants of the hits@k metric that vary by the number of positions in the ranking, are taken into account when calculating the metric (1 vs. 3 vs. 10).

Figure 7 shows the co-occurrence matrix for the ten most frequently used top-level classification metrics. *Accuracy* was most often reported together with *F-measure* metrics. Furthermore, *F-measure* metrics, *Precision* and *Recall* were frequently reported together.

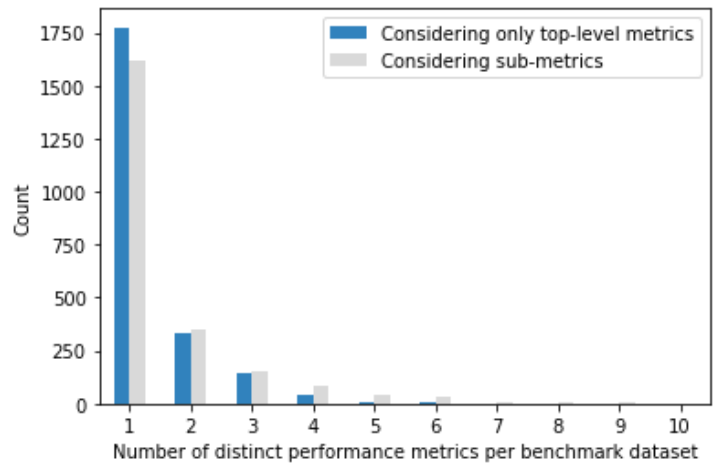

*Figure 6: Count of distinct metrics per benchmark dataset when considering only top-level metrics as distinct metrics (blue bars), and when considering sub-metrics as distinct metrics (grey bars).*

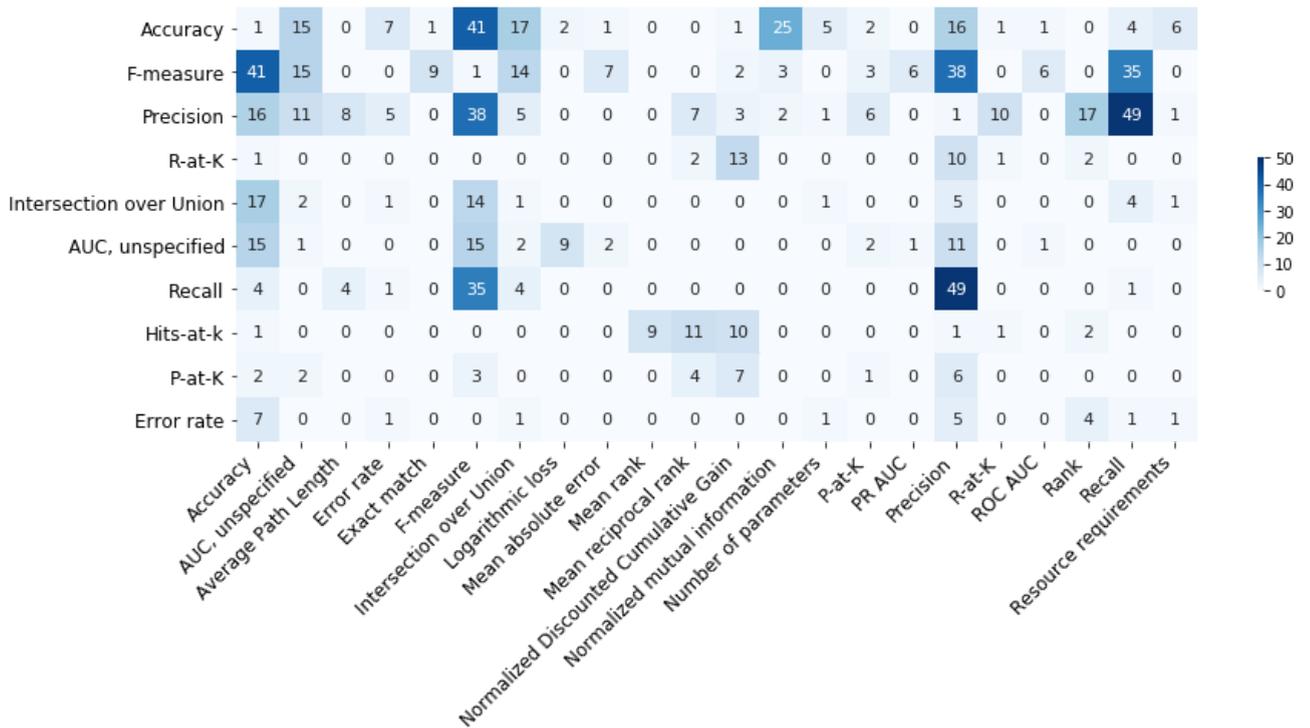

*Figure 7:* Co-occurrence matrix for the ten most frequently used top-level classification metrics (y-axis). Only top-level metrics that were reported at least five times together with either one of the selected top-level metrics are shown (x-axis). AUC, Area under the curve.

More complex metrics: the example of natural language processing

The three most commonly reported NLP-specific performance metrics were the "Bilingual Evaluation Understudy" (BLEU) score, the "Recall-Oriented Understudy for Gisting Evaluation" (ROUGE) metrics and the "Metric for Evaluation of Translation with Explicit ORdering" (METEOR), with BLEU score being by far the most frequently used metric (see Figure 5). Considering submetrics, ROUGE-1, ROUGE-2 and ROUGE-L were the most commonly annotated ROUGE variants, and BLEU-4 and BLEU-1 were the most frequently annotated BLEU variants. For a large fraction of BLEU and ROUGE annotations, the subvariant was not specified in the annotation.

The BLEU score was used across a wide range of NLP benchmark tasks, such as machine translation, question answering, summarization and text generation. ROUGE metrics were mostly used for text generation, video captioning and summarization tasks while METEOR was mainly used for image and video captioning, text generation and question answering tasks.

NLP-specific metrics occurring more rarely included Google-BLEU (GLEU) score, edit distance, phoneme and diacritic error rate and NIST, which were used to report model performance in less than three distinct benchmark datasets each.

The BLEU score was reported without any other metrics in 80.2% of the cases, whereas the ROUGE metrics more often appeared together with other metrics and stood alone in only nine out of 24 occurrences. METEOR was, in all cases, reported together with at least one other metric. Figure 8 shows the co-occurrence matrix for the top 10 most frequently used NLP-specific metrics. BLEU was most often reported together with the ROUGE metrics and METEOR, and vice versa.

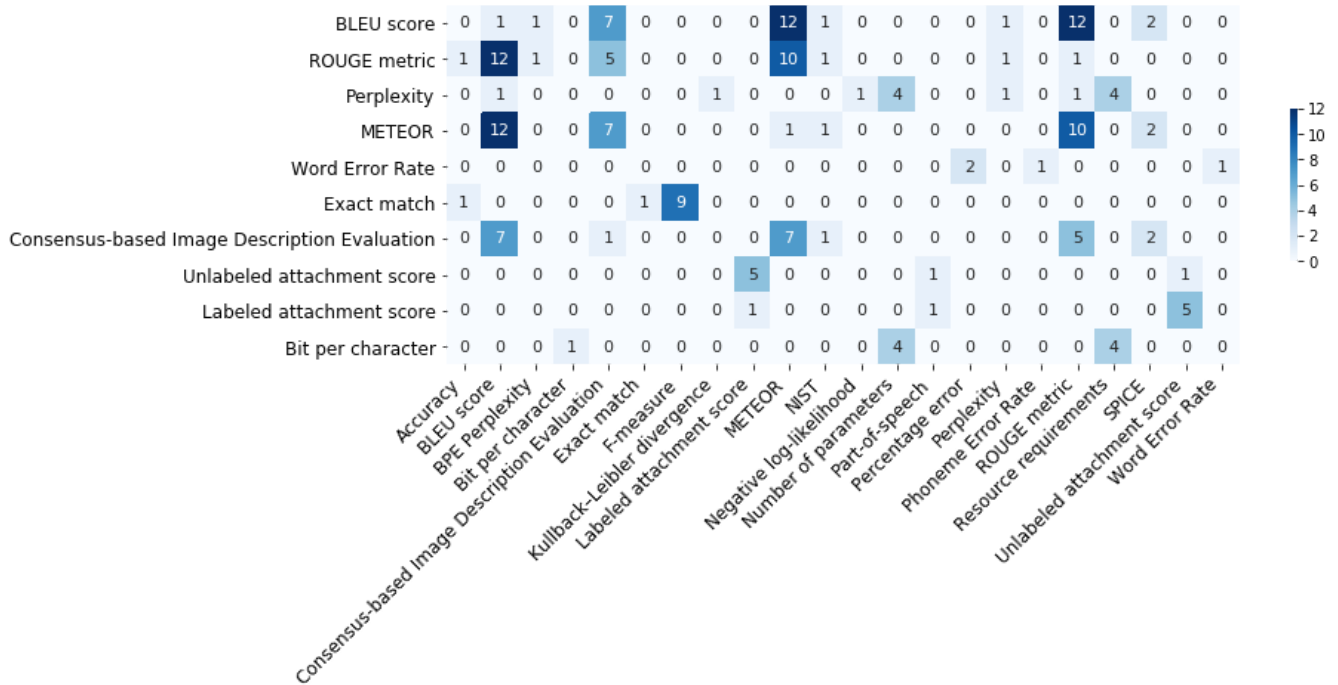

*Figure 8: Co-occurrence matrix for the top 10 most frequently used NLP metrics (y-axis). Only metrics that were reported at least one time together with either one of the selected metrics are shown (x-axis).*

## Inconsistencies and ambiguities in the reporting of performance metrics

During the mapping process it became evident that performance metrics are often reported in an inconsistent or ambiguous manner. For example, Area under the curve (AUC) metrics are often simply referred to as AUC, even though the curve and thus its interpretation can be a quite diverse, depending on whether it is drawn by plotting precision and recall against each other (PR-AUC), or recall and the false-positive rate (ROC-AUC), which are the two AUC metrics most commonly used by the Machine Learning community.

Another example are the ROUGE metrics, which have originally been proposed in different variants (e.g., ROUGE-1, ROUGE-L) but are often simply referred to as 'ROUGE'. Further, ROUGE metrics have originally been proposed in a 'recall' and 'precision' sub-variant, such as 'ROUGE-1 precision' and 'ROUGE-1 recall'. While it seems to be customary practice to assume that it is the recall variant because it is used most often, this might still lead to ambiguities when comparing results between different papers.

Similar inconsistencies and ambiguities exist for other performance metrics, such as weighted or macro and micro averages of scores, where a standardized definition may be lacking. In some cases, the metric used to report results could not be identified from the source paper.

# Discussion

## Metrics for evaluating classification tasks

*Accuracy, F1 score, precision* and *recall* were the most frequently used metrics to report model performance on benchmark datasets. As metrics for binary classification problems, they can be derived from a confusion matrix, a two by two contingency table of the predicted and observed class labels. Variants of these metrics for generalisation to multi-class classification, such as the 'micro' and 'macro' averages of the respective scores exist. (Sokolova and Lapalme 2009; Opitz and Burst 2019) Furthermore, several task-specific extensions of these scores have been conceptualized. For example, the relaxed F1 score takes into consideration inexact matches as well and has applications in Natural Language Processing tasks, such as Named Entity Recognition. Other NLP-specific metrics that can be seen as special variants of *precision* and *recall* include the BLEU, NIST, ROUGE and METEOR scores. Due to the added complexity of the language-component as compared to simple classification, they will be discussed separately.

Despite their exhaustive use in classification tasks, *accuracy, F1 score, precision* and *recall* exhibit a number of problematic properties that have subjected them to extensive criticism in the past decades.

*Accuracy* is the proportion of correct predictions (i.e. true positives and true negatives) to the overall number of observations. It can be used for evaluating binary and multiclass classifiers. Its major deficiency is its inability to yield informative results when dealing with unbalanced datasets, i.e. when there are large differences in the number of instances per class. This property may lead to the well-known phenomenon called 'accuracy paradox': If a classifier predicts the majority class in all cases, then accuracy is equal to the proportion of the majority class among the total cases. For example, if 'class A' makes up 95% of all instances, a classifier that predicts 'class A' all the time will have an accuracy of 95%.

*Precision* (or *positive predictive value*) is the proportion of true positives to the number of true positives and false positives. When interpreted as a probability, it estimates the probability that a randomly selected instance predicted as positive is a true positive. A classifier that yields no false positives has a precision of 1. *Recall* (or *sensitivity*) is the fraction of positive instances that are correctly classified as positive. When interpreted as a probability, it estimates the probability that a randomly selected true positive instance is predicted as positive. Precision focuses only on instances predicted as positive by the classifier while recall focuses only on the actual positive instances, classified as either positive or negative by the model. Both ignore the models' capacity to accurately predict negative cases.

The *F1 score* is a special case of the more general F-measure and is calculated as the harmonic mean of precision and recall, with both being given equal weight. The F-measure (or $F_\beta$) has a parameter β that specifies the balance between precision and recall and was derived from van Rijsbergen's effectiveness measure E. If β=1, the F1 score emerges with equal weight being given to precision and recall. (van Rijsbergen 1977) For β < 1 or β > 1, more weight is given to precision or recall, respectively. For example, $F_{0.5}$ gives twice as much weight to precision as to recall. While $F_\beta$ scores with a β different from 1 appeared in the analyzed dataset only once, the F1 score was one of the most widely used metrics for evaluating binary classifiers. Several shortcomings remain that have prompted researchers to reconsider its adequacy for evaluating classifiers on Machine Learning tasks. (Hand and Christen 2017; Chicco and Jurman 2020) Like accuracy, recall and precision, the F1 score may yield misleading results for classifiers that are biased towards predicting the majority class. Other potentially critical properties include its focus on only one class, its independence from the number of true negatives and its susceptibility to the swapping of class labels. (Powers 2015)

Finally, there are inconsistencies in the definition of F1 score for multi-class classification tasks. Opitz and Burst (2019) have found that two different formulas are currently in use to calculate macro F1, which only produce equivalent results under rare circumstances. (Sokolova and Lapalme 2009; Opitz and Burst 2019)

These shortcomings led to the proposal of a number of alternative confusion matrix-derived metrics. These include informedness and markedness (MK), Matthews Correlation Coefficient (MCC), Fowlkes–Mallows index (FM), macro average arithmetic (MAvA) and balanced accuracy. (Espíndola and Ebecken 2005; Ferri et al 2009; Brodersen et al 2010; Powers 2011; Chicco and Jurman 2020)

The Matthews Correlation Coefficient (MCC) is considered to be one of the most informative metrics by some researchers when dealing with imbalanced datasets. (Boughorbel et al 2017; Chicco and Jurman 2020) MCC is based on the entire confusion matrix and takes on values in the interval [-1, +1], where the values of -1 and +1 imply perfect misclassification and perfect classification, respectively, and 0 implies random predictions. MCC is also defined as the geometric mean of informedness and markedness, where informedness is the sum of the true positive and true negative rate minus 1, and markedness is the sum of positive and negative predictive value minus 1. (Powers 2011)

A comparative analysis of MCC, F1 score and accuracy across six different simulated scenarios can be found in (Chicco and Jurman 2020). They conclude that MCC most consistently provides an informative response across all scenarios, regardless of whether the dataset at hand is balanced (i.e., the number of instances per positive and negative class is roughly the same), positively imbalanced (i.e., the number of instances is much higher for the positive class) or negatively imbalanced (i.e., the number of instances is much higher for the negative class). In contrast, accuracy may lead to an overly optimistic assessment of a classifier's ability to generalize to new data in the cases of positively and negatively imbalanced datasets. The F1 score may have the same issue when dealing with positively imbalanced datasets, and balanced datasets where the classifier is biased towards predicting the positive class. In cases where a whole row or column of the confusion matrix is zero, MCC is undefined. It can, however, be assigned a value of zero, which matches the expected values in these constellations (Chicco and Jurman 2020). Despite its favourable properties, MCC was not used as a metric in any of the benchmark datasets included in our analysis.

The Fowlkes–Mallows index (FM) is defined as the geometric mean of precision and recall. It is sometimes, less specifically, referred to as Gmean or G-mean in the literature, perhaps because of this naming in several Machine learning libraries. (Fowlkes and Mallows 1983) However, while most sources treat Gmean or G-mean as equivalent to FM, they are sometimes also used to refer to the geometric mean of other metrics, e.g., recall and specificity. (Espíndola and Ebecken 2005) As FM takes into consideration and balances the classifiers' accuracy on both the positive and negative class, it is proposed as a potential alternative metric when dealing with imbalanced datasets. FM was not used as a performance metric by any of the benchmark datasets included in our analysis.

Other alternative classification metrics not appearing in our dataset include balanced accuracy, macro average arithmetic (MAvA), Cohen's κ coefficient, Cramér's V and K measure (a variant of informedness and balanced accuracy). (Sebastiani 2015; Brodersen et al. 2010; Ferri et al. 2009; Cohen 1960; Cramér 1946) Balanced accuracy is calculated by averaging over the fractions of correct predictions per class and can be used as an alternative to accuracy when dealing with an imbalanced dataset. For example, in case a classifier is biased towards the more frequent class, balanced accuracy would yield a lower result than accuracy. (Brodersen et al 2010) The macro average arithmetic (MAvA) is defined as the arithmetic average of the partial accuracies of each class. Rand index and adjusted rand index were used by a few datasets (2 and 4, respectively), however, not for classification but for segmentation and clustering tasks. (Hubert and Arabie 1985; Warrens 2008)

The classification metrics described above are all calculated based on a confusion matrix. Given that many models actually estimate class probabilities instead of class labels directly, the application of such metrics therefore requires a classification threshold. Another family of metrics for evaluating binary classifiers based on all potential thresholds exists. Area under the curve (AUC) metrics are based on the curve resulting from the comparison of two confusion matrix-derived metrics for all possible states of a specific classifiers' confusion matrix defined by all potential decision thresholds. The most commonly used AUC metrics are Receiver Operator Characteristic AUC (ROC-AUC, C-statistic, C-index) and Precision-Recall AUC (PR-AUC). ROC-AUC, PR-AUC and not further specified AUC metrics were among the ten most common classification metrics in our analyzed dataset. ROC-AUC is defined by the trade-off between the true positive rate (recall, sensitivity) and the false positive rate. PR-AUC is defined by the tradeoff between the true positive rate (recall) and positive predictive value (precision). Both ROC-AUC and PR-AUC take on values ranging from 0 to 1 but typical values range from 0.5 (corresponding to random guessing) and 1 (corresponding to perfect classification). While ROC-AUC can be interpreted as the probability that the predicted risk is higher for a randomly selected case than a randomly selected non-case, PR-AUC lacks a similar intuitive interpretation.

AUC metrics have been proposed as alternatives to accuracy and F1 score when dealing with small or unbalanced datasets, with the PR curve, as compared to the ROC curve, having been discussed as the more informative metric for imbalanced datasets. (Davis and Goadrich 2006; Ferri et al 2009; Chicco and Jurman 2020) Davis and Goadrich have investigated the relationship of ROC curves and PR curves and found that algorithms that optimize ROC-AUC do not necessarily optimize PR-AUC. (Davis and Goadrich 2006)

Furthermore, cost curves have frequently been proposed as an alternative to ROC curves when dealing with imbalanced datasets. (Drummond and Holte 2004; Chawla et al 2004) Variants of the AUC for multi-class classification, such as AUNU or AUNP, exist but seem to be used to a much lesser extent by the Machine Learning community. (Fawcett 2001; Fawcett 2006; Ferri et al 2009) Moreover, pinned AUC was proposed as a variant of ROC-AUC for measuring unintended bias in classification settings with class imbalances. It has, however, later been shown to be susceptible to provide a skewed measurement of bias. (Dixon et al 2018; Borkan et al 2019)

## Metrics for evaluating natural language processing tasks

Natural Language Processing (NLP) is a very broad field that covers a wide range of different tasks and thus shows a large diversity in terms of the metrics that are used for performance comparison on benchmark datasets. For tasks that are generally treated as classification tasks, e.g., named entity recognition and part-of-speech tagging, associated metrics have been discussed in the above section 'Metrics for evaluating classification tasks'. Other, more complex tasks that require different evaluation metrics include machine translation, question answering, and summarization. Metrics designed for these tasks generally aim to assess the similarity between a machine-generated text and a reference text or set of reference texts that are human-generated.

BLEU score, ROUGE metrics and METEOR (see Figure 5) were the most frequently used metrics for evaluating NLP tasks that require evaluating text fragments based on reference texts.

BLEU (Bilingual Evaluation Understudy Score) score was proposed by IBM in 2001 as an automatic metric for Machine translation tasks. (Papineni et al 2001) It is based on n-gram precision (i.e., the fraction of matching n-grams to the total number of generated n-grams) and applies a so-called 'brevity penalty', which penalizes translations that show significantly shorter length compared to the reference translation. While the original BLEU score was not designed for sentence-level comparison due to its geometric averaging of n-grams, a variant called 'Smoothed BLEU' (BLEUS) for sentence-level comparison was later proposed to address this issue. (Lin and Och 2004)

The BLEU score continues to be one of the most frequently used metrics for machine translation and other language-generating tasks. However, several weaknesses have been pointed out by the research community, such as its sole focus on n-gram precision without considering recall and its reliance on exact n-gram matchings. Zhang et al. have discussed properties of the BLEU score and NIST, a variant of the BLEU score that gives more weight to rarer n-grams than to more frequent ones, and came to the conclusion that neither of the two metrics necessarily show high correlation with human judgements of machine translation quality. (Doddington 2002; Zhang et al 2004)

METEOR (Metric for Evaluation of Translation with Explicit Ordering) was proposed in 2005 to address weaknesses of previous metrics. (Banerjee and Lavie 2005) METEOR is an F-measure derived metric that has repeatedly been shown to yield higher correlation with human judgment across several tasks as compared to BLEU and NIST. (Lavie et al 2004; Graham et al 2015; Chen et al 2019). Matchings are scored based on their unigram precision, unigram recall (given higher weight than precision), and a comparison of the word ordering of the translation compared to the reference text. This is in contrast to the BLEU score, which does not take into account n-gram recall. Furthermore, while BLEU only considers exact word matches in its scoring, METEOR also takes into account words that are morphologically related or synonymous to each other by using stemming, lexical resources and a paraphrase table. Additionally, METEOR was designed to provide informative scores at sentence-level and not only at corpus-level.

An adapted version of METEOR, called METEOR++ 2.0, was proposed in 2019. (Guo and Hu 2019) This variant extends METEOR's paraphrasing table with a large external paraphrase database and has been shown to correlate better with human judgement across many machine translation tasks.

Compared to BLEU and ROUGE, METEOR was rarely used as a performance metric (n=13) across the NLP benchmark datasets included in our dataset.

The ROUGE (Recall-Oriented Understudy for Gisting Evaluation) metrics family was the second most used NLP-specific metric in our dataset after the BLEU score. It was introduced in 2004 as a set of metrics to evaluate machine-generated summaries based on reference texts. The original ROUGE set proposes metrics for measuring the overlap of n-grams between generated and reference summaries (ROUGE-N), the longest shared sequence (ROUGE-L), the longest shared sequence while taking word-order into account (ROUGE-W), skip-bigram co-occurrence (ROUGE-S) and ROUGE-SU, which combines skip-bigram with unigram counts. (Lin 2004) In general, all ROUGE metrics can be calculated in terms of precision (i.e., how many of the n-grams in the machine-generated text appear in the reference text) and recall (i.e., how many of the n-grams in the re text appear in the machine-generated text), and the resulting F1 score can be calculated.

While originally proposed for summarization tasks, a subset of the ROUGE metrics (i.e. ROUGE-L, ROUGE-W and ROUGE-S) has also been shown to perform well in machine translation evaluation tasks. (Lin 2004; Och 2004) However, the ROUGE metrics set has also been shown to not adequately cover multi-document summarization, tasks that rely on extensive paraphrasing, such as abstractive summarization, and extractive summarization of multi-logue text types (i.e. transcripts with many different speakers), such as meeting transcripts. (Lin 2004; Liu and Liu 2008; Ng and Abrecht 2015) Several new variants have been proposed in recent years, which make use of the incorporation of word embeddings (ROUGE-WE), graph-based approaches (ROUGE-G), or the extension with additional lexical features (ROUGE 2.0) (Ng and Abrecht 2015; ShafieiBavani et al 2018; Ganesan 2018)) ROUGE-1, ROUGE-2 and ROUGE-L were the most common ROUGE metrics used as performance metrics in our analyzed dataset.

The GLEU score was proposed as an evaluation metric for NLP applications, such as machine translation, summarization and natural language generation, in 2007. (Mutton et al 2007) It is a Support Vector Machine-based metric that uses a combination of individual parser-derived metrics as features. GLEU aims to assess how well the generated text conforms to 'normal' use of human language, i.e., its 'fluency'. This is in contrast to other commonly used metrics that focus on how well a generated text reflects a reference text or vice versa. GLEU appeared 3 times as a performance metric in our dataset.

Alternative NLP-specific metrics that have been proposed by the NLP research community but do not appear as performance metrics in the analyzed dataset include Translation error rate (TER), TER-Plus, "Length Penalty, Precision, n-gram Position difference Penalty and Recall" (LEPOR), Sentence Mover's Similarity, and BERTScore.

TER was proposed as a metric for evaluating machine translation quality. TER measures quality by the number of edits that are needed to change the machine-generated text into the reference text(s), with lower TER scores indicating higher translation quality. (Makhoul 2006) TER considers five edit operations to change the output into the reference text: Matches, insertions, deletions, substitutions and shifts. An adaptation of TER, TER-Plus, was proposed in 2009. Ter-Plus extends TER with three additional edit operations, i.e., stem matches, synonym matches and phrase substitution. (Snover et al 2009) TER-Plus was shown to have higher correlations with human judgements in machine translation tasks than BLEU, METEOR and TERp (Snover et al 2009)

LEPOR and its variants hLEPOR and nLEPOR were proposed as a language-independent model that aims to address the issue that several previous metrics tend to perform worse on languages other than those it was originally designed for. It has been shown to yield higher correlations with human judgement than METEOR, BLEU, or TER. (Han et al 2012)

Sentence Mover's Similarity (SMS) is a metric based on ELMo word embeddings and Earth mover's distance, which measures the minimum cost of turning a set of machine generated sentences into a reference text's sentences. (Peters et al 2018; Clark et al 2019) It was proposed in 2019 and was shown to yield better results as compared to ROUGE-L in terms of correlation with human judgment in summarization tasks.

BERTScore was proposed as a task-agnostic performance metric in 2019. (Zhang et al 2019) It computes the similarity of two sentences based on the sum of cosine similarities between their token's contextual embeddings (BERT), and optionally weighs them by inverse document frequency scores (Devlin et al 2018) BERTScore was shown to outperform established metrics, such as BLEU, METEOR and ROUGE-L in machine translation and image captioning tasks. It was also more robust than other metrics when applied to an adversarial paraphrase detection task. However, the authors also state that BERTScore's configuration should be adapted to task-specific needs since no single configuration consistently outperforms all others across tasks.

Due to the various shortcomings of currently used automatic evaluation metrics, metric development for language-generation tasks is an open research question. Metric evaluation was even introduced as an independent task at the annual Machine Translation conference. (Ma et al 2019)

Difficulties associated with automatic evaluation of machine generated texts include poor correlation with human judgement, language bias (i.e. the metric shows better correlation with human judgment for certain languages than others), and worse suitability for language generation tasks other than the one it was proposed for. (Novikova et al 2017) In fact, most NLP metrics have originally been conceptualized for a very specific application, such as BLEU and METEOR for machine translation, or ROUGE for the evaluation of machine generated text summaries, but have since then been introduced as metrics for several other NLP tasks, such as question-answering, where all three of the above mentioned scores are regularly used. Non-transferability to other tasks has recently been shown by Chen et al. who have compared several metrics (i.e. ROUGE-L, METEOR, BERTScore, BLEU-1, BLEU-4, Conditional BERTScore and Sentence Mover's Similarity) for evaluating generative Question-Answering (QA) tasks based on three QA datasets. They recommend that from the evaluated metrics, METEOR should preferably be used and point out that metrics originally introduced for evaluating machine translation and summarization do not necessarily perform well in the evaluation of question answering tasks. (Chen et al 2019)

Comparative evaluation studies currently seem to consider only a small set of metrics and are focused on specific NLP-tasks, while large comparative studies across multiple tasks are, to the best of our knowledge, yet to be undertaken.

Finally, many NLP metrics use very specific sets of features, such as specific word embeddings or linguistic elements, which may complicate comparability and replicability, which is of especially high relevance when considering the current process of measuring progress in terms of standardized performance metrics on benchmark datasets. To address the issue of replicability, reference open source implementations have been published for some metrics, such as, ROUGE, sentBleu-moses as part of the Moses toolkit and sacreBLEU. (Lin 2004)

### Limitations

The results presented in this paper are based on a large set of Machine learning papers available from the PWC database, which is – to the best of our knowledge – the currently largest available annotated dataset. The database comprises both preprints of papers published on arXiv and papers published in peer-reviewed journals. While it could be argued that arXiv preprints are not representative of scientific journal articles, it has recently been shown that a large fraction of arXiv preprints (77%) are subsequently published in peer-reviewed venues. (Lin et al 2020)

The descriptive statistics we present reflect performance metric annotations on PWC, which might not be an exact representation of the frequency of use in the original papers. In some cases, additional results might be available in the source papers. In these cases, the annotated metric reflects the annotator's selection.

In our analysis, we focused on classification metrics and, as an example for more complex metrics, on performance metrics used to evaluate NLP-specific tasks. We did not discuss performance metrics for point-estimation tasks. Metrics for regression tasks, such as mean squared error (MSE), mean absolute error (MAE), root mean square deviation (RMSD) and $R^2$ appeared rarely in the analyzed dataset, being used by only 5% of benchmark datasets. Nevertheless, these metrics have also been extensively discussed in the literature. (Armstrong and Collopy 1992; Harrell 2001; Chai and Draxler 2014; Botchkarev 2019) Finally, we did not discuss classification metrics that measure deviations using probabilities including proper scoring rules.

# Conclusions

The large majority of metrics currently used to evaluate classification AI benchmark tasks have properties that may result in an inadequate reflection of a classifiers' performance, especially when used with imbalanced datasets. While alternative metrics that address problematic properties have been proposed, they are currently rarely applied as performance metrics in benchmarking tasks, where a small set of historically established metrics is used instead. NLP-specific tasks pose additional challenges for metrics design due to language and task-specific complexities. Large comparative evaluation studies of different NLP-specific metrics across multiple benchmarking tasks are yet to be performed. Finally, we noticed that the reporting of metrics was partly inconsistent and partly unspecific, which may lead to ambiguities when comparing model performances.

# Methods

### Raw dataset

Our analyses are based on the data available from Papers with Code[3] (PWC), a large web-based open platform that collects Machine learning papers including code implementation of models and summarizes evaluation results on benchmark datasets, allowing a quick overview of current state-of-the-art results. PWC data is curated by combining automatic extraction from arXiv submissions and manual crowd-sourced annotation of results.

### Intelligence Task Ontology (ITO)

The Intelligence Task Ontology (ITO)[4] is an ontology that aims to provide a comprehensive map of artificial intelligence tasks using a richly structured hierarchy of processes, algorithms, data and performance metrics. ITO is based on data from PWC and the EDAM ontology[5]. The creation of ITO will be described in another manuscript.

We used ITO for further curation and creation of a hierarchical mapping of the raw performance metric data from PWC.

### Hierarchical mapping and further curation of metric names

The raw dataset exported from PWC contained a total number of 812 different strings representing metric names that appeared as distinct data property instances in ITO. These metric names were used by human annotators on the PWC platform to add results for a given model to the evaluation table of the relevant benchmark dataset's leaderboard on PWC.

This list of raw metrics in the PWC database was manually curated into a canonical hierarchy by our team. This entailed some complexities and required extensive manual curation which was conducted based on the following mapping procedure:

In many cases, the same metric was reported under many different synonyms and abbreviations. For example, for the F1 score, 63 naming variations appeared throughout the dataset, such as 'H-Mean', 'F1' and 'f1-score'. These were made subproperties of the property

---

[3] https://paperswithcode.com/
[4] https://github.com/OpenBioLink/ITO
[5] http://edamontology.org/

'F1 score'. 'F1 score' was further mapped to the top-level metric 'F-Measure', which denotes its general form.

In case the same metric appeared under different well-established designations in the dataset, the name that was more commonly used in Machine Learning literature was selected as the property name, and alternative names were mapped to this name. For example, 'Sensitivity' was made a sub-property of 'Recall' and Sørensen–Dice coefficient was subsumed under F1 score.

Some performance metrics were reported in different sub-variants. For example, the performance metrics 'hits@1', 'hits@3' and 'hits@10' were made sub-properties of hits@k.

Metric names in PWC that contained task-specific modifications, such as 'Q8 accuracy' or 'Laptop (F1)', were mapped to the respective top-level metrics, in this case 'Accuracy' and 'F1 score', respectively.

Some annotators used the name of a benchmark dataset, such as 'SICK-E' or 'Daily mail' instead of the actual metric name. In these cases, the performance metrics that were used with this benchmark dataset were manually looked up from the respective papers and the properties were assigned accordingly. In some cases, this required assigning a property to multiple super-properties, since more than one metric was associated with the benchmark dataset.

In case a library that implemented a metric was used as the metric name, e.g., SacreBLEU, which is a reference implementation of the BLEU score available as a Python package, this property was made sub-metric of the more general metric name, in this case 'BLEU score'.

For some metric names included in the raw data export, the metric was not identifiable from the string, e.g. 'Kitchen', 'Sentiment' or 'Books'. In these cases, we manually obtained the metric names from the respective source paper and assigned them accordingly.

271 entries from the original list, could not be assigned a metric and were subsumed under a separate category 'Undetermined'. After this extensive manual curation, the resulting list covered by our dataset could be reduced from 812 to 187 distinct performance metrics.

Whenever possible and sensible, we used the respective preferred Wikipedia article titles as canonical names for the metrics.

## Grouping of top-level metrics

Top-level metrics were further grouped into categories based on the task type they are usually applied to: Classification, Computer vision, Natural language processing, Regression, Game Playing, Ranking, Clustering and 'Other'. The categories 'Computer vision' and 'Natural language processing' contain only metrics that are specific to these two task types, e.g., 'Inception score' or 'METEOR', respectively.

## Analysis

Analyses were performed based on the ITO version of 13.7.2020. Raw statistics were generated based on the ITO ontology using SPARQL queries and further processed and analyzed using Jupyter Notebooks and the Python 'pandas' library. Data, code and notebooks to generate these statistics are available on Github (see section 'Data and code availability').

# Abbreviations

| | |
|---|---|
| AUC | Area under the curve |
| BLEU | Bilingual Evaluation Understudy Score |
| FM | Fowlkes–Mallows index |
| ITO | Intelligence task ontology |
| MAE | Mean absolute error |
| MAvA | Macro average arithmetic |
| MCC | Matthews Correlation Coefficient |
| METEOR | Metric for Evaluation of Translation with Explicit Ordering |
| MK | Markedness |
| MSE | Mean squared error |
| NIST | National Institute of Standards and Technology |
| NLP | Natural language processing |
| PR-AUC | Precision recall area under the curve |
| PWC | Papers with code |
| RMSD | Root mean square deviation |
| ROC-AUC | Receiver Operator Characteristic area under the curve |
| ROUGE | Recall-Oriented Understudy for Gisting Evaluation |
| SMS | Sentence Mover's Similarity |
| TER | Translation error rate |

# Declarations

## Funding

A part of the research leading to these results has received funding from the European Community's Horizon 2020 Programme under grant agreement No. 668353 (U-PGx).

## Competing Interests

The authors declare no competing interests.

## Data and code availability

The OWL (Web Ontology Language) file of the ITO model is made available on Github[6] and BioPortal[7]. The ontology file is distributed under a CC-BY-SA license. ITO includes data from the Papers With Code project (https://paperswithcode.com/). Papers With Code is licensed under the CC-BY-SA license. Data from Papers With Code are partially altered (manual curation to improve ontological structure and data quality). ITO includes data from the EDAM ontology. The EDAM ontology is licensed under a CC-BY-SA license.

Notebooks containing the SPARQL queries to generate statistics from ITO and Python code for analysing the data are also accessible via GitHub[2]. Furthermore, the repository includes complete statistics for all performed analyses and a mapping of commonly used naming variants of the most frequently used performance metrics. Additionally, we provide an overview of the basic properties, e.g., score ranges and whether higher or lower scores are better, of the top 50 most frequently used performance metrics.

## Author Contributions

**Kathrin Blagec**: Conceptualization, Data curation, Methodology, Formal analysis, Investigation, Visualization, Writing - Original Draft. **Georg Dorffner:** Writing - Review & Editing. **Milad Moradi:** Writing - Review & Editing. **Matthias Samwald:** Conceptualization, Data curation, Methodology, Formal analysis, Writing - Review & Editing, Funding acquisition.

## Acknowledgements

We thank the team from 'Papers With Code' for making their database available and all annotators who contributed to it.

Furthermore, we thank Michael Kammer for valuable comments and discussions.

---

[6] https://github.com/OpenBioLink/ITO
[7] https://bioportal.bioontology.org/ontologies/ITO